\documentclass[11pt,a4paper]{article}
\usepackage[hyperref]{naaclhlt2019}
\usepackage{times}
\usepackage{latexsym}

\usepackage{url}
\usepackage{amsmath,amssymb}
\DeclareMathOperator{\E}{\mathbb{E}}

\usepackage{amsfonts}
\usepackage{helvet}
\usepackage{courier}
\usepackage{url}
\usepackage{float,color}
\usepackage{times}
\usepackage{algorithm}
\usepackage{algorithmic}
\usepackage{epsfig}
\usepackage{stfloats}
\usepackage{url}
\usepackage{epstopdf}
\usepackage{multicol}
\usepackage{tocstyle}
\usepackage{dsfont,amsfonts,amssymb,amsmath,color}
\usepackage{multirow}
\usepackage{booktabs}
\def\0{{\bf 0}}
\def\1{{\bf 1}}

\def\LM{{\mathcal L}}

\def\NM{{\mathcal N}}

\aclfinalcopy 

\title{A Variational Approach to Weakly Supervised Document-Level Multi-Aspect Sentiment Classification}

\author{Ziqian Zeng\textsuperscript{1}, Wenxuan Zhou\textsuperscript{2}, Xin Liu\textsuperscript{1}, and Yangqiu Song\textsuperscript{1} \\
  \textsuperscript{1}Department of CSE, Hong Kong University of Science and Technology, HK\\
  \textsuperscript{2}Department of CS, University of Southern California, CA, USA \\
   {\tt \textsuperscript{1}\{zzengae, xliucr, yqsong\}@cse.ust.hk } \\
    {\tt \textsuperscript{2}\{zhouwenx\}@usc.edu}\\}
\date{}

\begin{document}
\maketitle

\begin{abstract}
In this paper, we propose a variational approach to weakly supervised document-level multi-aspect sentiment classification.
Instead of using user-generated ratings or annotations provided by domain experts, we use target-opinion word pairs as ``supervision.'' 
These word pairs can be extracted by using dependency parsers and simple rules. 
Our objective is to predict an opinion word given a target word while our ultimate goal is to learn a sentiment polarity classifier to predict the sentiment polarity of each aspect given a document. 
By introducing a latent variable, i.e., the sentiment polarity, to the objective function, we can inject the sentiment polarity classifier to the objective via the variational lower bound. 
We can learn a sentiment polarity classifier by optimizing the lower bound.
We show that our method can outperform weakly supervised baselines on TripAdvisor and BeerAdvocate datasets and can be comparable to the state-of-the-art supervised method with hundreds of labels per aspect.
\end{abstract} 
\section{Introduction}
Document-level multi-aspect sentiment classification (DMSC) aims to predict the sentiment polarity of each aspect given a document which consists of several sentences describing one or more aspects~\cite{wang2010latent,wang2011latent,yin2017document}.
Solving the DMSC task is useful for providing both recommendations for users and suggestions for business owners on customer review platforms.

Aspect based sentiment classification~\cite{tang2016effective, tang2016aspect, wang2016attention, chen2017recurrent, ma2017interactive, wang2018aspect} was usually done by supervised learning, where aspect-level annotations should be provided. 
Aspect-level annotations are not easy to obtain. 
Even when the platform provides the function to rate for different aspects, users are less likely to submit all of them. 
For example, {about 37\%} of the aspect ratings are missing on \href{https://www.tripadvisor.com.hk/}{TripAdvisor}. 
If we can solve DMSC task without using aspect-level annotations, it can save human effort to annotate data or collect user-generated annotations on the platform. 

Existing weakly supervised approaches~\cite{wang2010latent,wang2011latent} use overall polarities instead of aspect polarities as ``supervision.'' 
Compared with the polarity of each aspect, it is relatively easy to obtain overall polarities. 
Specifically, they minimize the square loss between the overall polarity and the weighted sum of all aspect polarities.
However, when users only care about a particular rare aspect, e.g., childcare services, these approaches cannot estimate parameters of the rare aspect incrementally. 
They have to re-collect documents which mentioned this rare aspect and estimate parameters of all aspects based on the new corpus.  
In addition, these approaches assume the document is a bag-of-words, which neglects the order of the words and fails to capture the similarity between words.

In this paper, we propose to use target-opinion word pairs as ``supervision.'' 
Target-opinion word pairs can be helpful with our ultimate goal which is to learn a classifier to predict the sentiment polarity of each aspect given a document.  
For example, in a document ``The bedroom is very spacious,'' if we can extract the target-opinion pair ``bedroom-spacious,'' the sentiment polarity of the aspect \textit{room} is likely to be \textit{positive}. 
Hence, we propose to achieve the polarity classification goal by accomplishing another relevant objective: to predict an opinion word given a target word. 

We can decompose the opinion word prediction objective into two sub-tasks.
The first sub-task is to predict the sentiment polarity based on a document.
For example, given a document ``The bedroom is very spacious,'' it predicts the sentiment polarity of the aspect \textit{room} to be \textit{positive}. 
The second sub-task is to predict the opinion word given a target word and a sentiment polarity predicted by the first sub-task.
For example, knowing the fact that the sentiment polarity of the aspect \textit{room} is \textit{positive}, it predicts the opinion word associated with the target word ``room'' to be ``spacious.'' 
By introducing a latent variable, i.e., the sentiment polarity of an aspect, to the opinion word prediction objective, we can inject the polarity classification goal (the first sub-task) into the objective via the variational lower bound which also incorporates the second sub-task. 
In this sense, our training objective is only based on the target-opinion word pairs which can be extracted by using dependency parsers and some manually designed rules. 
We consider our approach as weakly supervised learning because there is no direct supervision from polarity of each aspect.

In other words, our model includes two classifiers: a sentiment polarity classifier and an opinion word classifier.
In the sentiment polarity classifier, it predicts the sentiment polarity given a document.
In the opinion word classifier, it predicts an opinion word based on a target word and a sentiment polarity. 
Compared with previous approaches~\cite{wang2010latent,wang2011latent}, our approach can get rid of the assumption that the overall polarity should be observed and it is a weighted sum of all aspect polarities. 
Moreover, our approach can estimate parameters of a new aspect incrementally. 
In addition, our sentiment polarity classifier can be more flexible to capture dependencies among words beyond the bag-of-words representation if we use a deep neural network architecture to extract features to represent a document. 
We conducted experiments on two datasets, TripAdvisor~\cite{wang2010latent} and BeerAdvocate~\cite{mcauley2012learning}, to illustrate the effectiveness of our approach.

Our contributions are summarized as follows,

$\bullet$ We propose to solve DMSC task in a nearly unsupervised way.

$\bullet$ We propose to learn a classifier by injecting it into another relevant objective via the variational lower bound. This framework is flexible to incorporate different kinds of document representations and relevant objectives.

$\bullet$ We show promising results on two real datasets and we can produce comparable results to the supervised method with hundreds of labels per aspect.

Code and data for this paper are available on \url{https://github.com/HKUST-KnowComp/VWS-DMSC}.
\begin{figure}
    \center
    \includegraphics[width=0.38\textwidth]{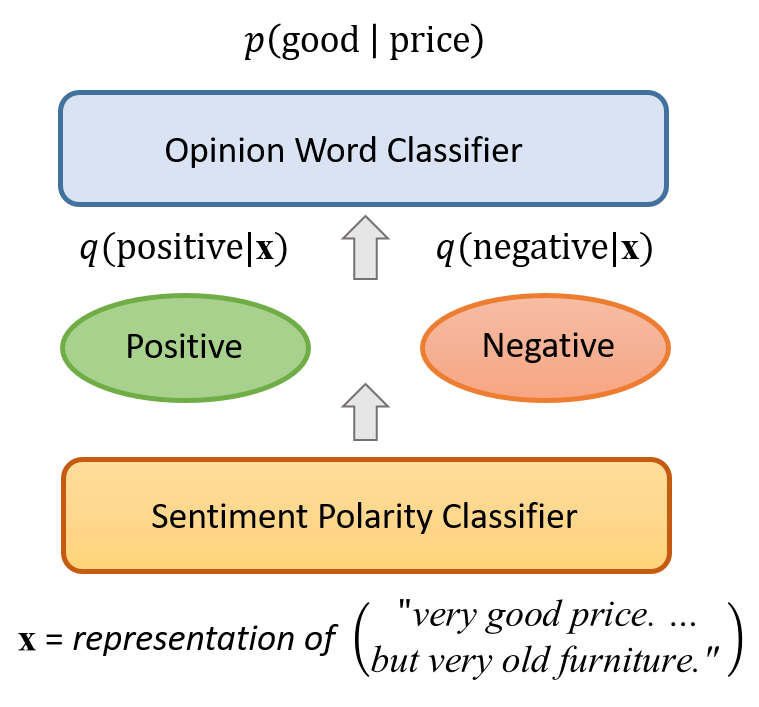}\vspace{-0.05in}
    \caption{A sentiment polarity classifier and an opinion word classifier associated with the aspect \textit{price}.}\label{fig:model-illustrate}
\end{figure}

\section{VWS-DMSC Approach}
In this section, we describe our variational approach to weakly supervised DMSC (VWS-DMSC).
In the next section, we present how we obtain target-opinion word pairs by using a rule-based extraction approach.

\subsection{Overview} 
Our model consists of a sentiment polarity classifier and an opinion word classifier. Our task is document-level multi-aspect sentiment classification. For each aspect, we train a sentiment polarity classifier and an opinion word classifier. The input of the sentiment polarity classifier of each aspect is the same, i.e., a representation of a document. The target-opinion word pairs used in opinion word classifiers are different for different aspects. 

Figure \ref{fig:model-illustrate} shows the relation between two classifiers (on the aspect \textit{price}). 
The input $\mathbf{x}$ of the sentiment polarity classifier is a representation of a document, e.g., bag-of-words or a representation learned by recurrent neural networks.
The sentiment polarity classifier takes $\mathbf{x}$ as input and produces a distribution of sentiment polarity $R_a$ of an aspect $a$, denoted as $q(R_a|\mathbf{x})$. 
If $R_a$ only has two possible values, i.e., positive and negative, then outputs of the classifier are $q(positive|\mathbf{x})$ and $q(negative|\mathbf{x})$ respectively.
The opinion word classifier takes a target word (``price'') and a possible value of the sentiment polarity $r_a$ as input, and estimates $p(\text{``good''}|r_a,\text{``price''})$. 
Our training objective is to maximize the log-likelihood of an opinion word given a target word, e.g., $p(\text{``good''}|\text{``price''})$. The likelihood is estimated based on the sentiment polarity classifier and the opinion word classifier.

\subsection{Sentiment Polarity Classifier}
The sentiment polarity classifier aims to estimate a distribution of sentiment polarity $q(R_a|\mathbf{x})$, where $R_a$ is a discrete random variable representing the sentiment polarity and $\mathbf{x}$ is a feature representation of a document. 
We use a simple Softmax classifier here. 
We denote $r_a$ as a possible value of the random variable $R_a$, representing a possible sentiment polarity.
The model estimates the probability of class $r_a$ as
\begin{equation}\label{eq:encoder-softmax}
q(R_a = r_a|\mathbf{x}) = \frac{\exp \big( \mathbf{w}_{r_a}^{T} \mathbf{x} \big)}{\sum_{r_a'}{\exp \big( \mathbf{w}_{r_a'}^{T} \mathbf{x} \big)}} \; ,
\end{equation}
where $\mathbf{w}_{r_a}$ is a vector associated with sentiment class $r_a$ for aspect $a$.

\paragraph{Document Representation}
The representation of a document $\mathbf{x}$ can be different using different feature extraction approaches. 
Traditional document representations of sentiment classification would be bag-of-words, n-gram, or averaged word embeddings. 
Recently, end-to-end recurrent neural network based models demonstrate a powerful capacity to extract features of a document. 
The state-of-the-art model in DMSC task is \cite{yin2017document}. 
We use it as the document representation in our model.
\subsection{Opinion Word Classifier}
The opinion word classifier aims to estimate the probability of an opinion word $w_o$ given a target word $w_t$ and a sentiment polarity $r_a$:
\begin{equation}\label{sp}
p(w_o |r_a, w_t) = \frac{\exp \big( \varphi ({w_o}, {w_t}, r_a) \big)}{ \sum_{{w_o'}}{ \exp \big( \varphi ({w_o'}, {w_t}, r_a) \big) } } \; ,
\end{equation}
where $\varphi$ is a scoring function related to opinion word $w_o$, target word $w_t$, and sentiment polarity $r_a$. 
Here we use the dot product as the scoring function:
\begin{equation}
\varphi ({w_o}, {w_t}, r_a)= I \big( (w_t,w_o) \in \mathcal{P}, w_t \in \mathcal{K}_a \big )\cdot \mathbf{c}_{r_a}^T \mathbf{w}_o \;,
\end{equation}
where $\mathbf{w}_o$ is the word embedding of opinion word $w_o$, $\mathbf{c}_{r_a}$ is a vector associated with $r_a$, $\mathcal{P}$ is the set of pairs extracted from the document, $\mathcal{K}_a$ is the set of target words associated with aspect $a$, and
$I(\cdot)$ is an indicator function where $I(true)=1$ and $I(false)=0$.

Given a target word $w_t$ and a sentiment polarity $r_a$, we aim to maximize the probability of opinion words highly related to them.
For example, opinion word ``good'' is usually related to target word ``price'' for aspect \textit{value} with sentiment polarity \textit{positive}, and opinion word ``terrible'' is usually related to target word ``traffic'' for aspect \textit{location} with sentiment polarity \textit{negative}.

\subsection{Training Objective}
The objective function is to maximize the log-likelihood of an opinion word $w_o$ given a target word $w_t$. 
As we mentioned before, the objective function can be decomposed into two sub-tasks.
The first one corresponds to the sentiment polarity classifier. 
The second one corresponds to the opinion word classifier. 
After introducing a latent variable, i.e., the sentiment polarity, to the objective function, we can derive a variational lower bound of the log-likelihood which can incorporate two classifiers:

\begin{align}
\LM & = \log p(w_o | w_t) \nonumber \\
& = \log \sum_{r_a}p(w_o , r_a | w_t) \nonumber\\
& = \log \sum_{r_a} q(r_a|\mathbf{x}) \Big[ \frac{p(w_o , r_a | w_t)}{q(r_a|\mathbf{x})} \Big] \nonumber \\
& \geq \sum_{r_a} q(r_a|\mathbf{x}) \Big[ \log \frac{p(w_o , r_a | w_t)}{q(r_a|\mathbf{x})} \Big] \nonumber \\
& = \E_{q(R_a|\mathbf{x})} \big[\log p(w_o|r_a,w_t)p(r_a|w_t) \big] + H(q(R_a|\mathbf{x})) \nonumber \\
& = \E_{q(R_a|\mathbf{x})} \big[\log p(w_o|r_a,w_t)p(r_a) \big] + H(q(R_a|\mathbf{x})) \;, \label{vari}
\end{align}
where $H(\cdot)$ refers to the Shannon entropy.
By applying Jensen's inequality, the log-likelihood is lower-bounded by Eq. (\ref{vari}). 
The equality holds if and only if the KL-divergence of two distributions, $q(R_a|\mathbf{x})$ and $p(R_a|w_t,w_o)$, equals to zero. 
Maximizing the variational lower bound is equivalent to minimizing the KL-divergence. 
Hence, we can learn a sentiment polarity classifier which can produce a similar distribution to the true posterior $p(R_a|w_t,w_o)$. 
Compared with $p(R_a|w_t,w_o)$, $q(R_a|\mathbf{x})$ is more flexible since it can take any kind of feature representations as input. 
We assume that a target word $w_t$ and a sentiment polarity $r_a$ are independent since the polarity assignment is not influenced by the target word. 
We also assume that the sentiment polarity $R_a$ follows a uniform distribution, which means $p(r_a)$ is a constant. 
We remove it in Eq. (\ref{vari}) to get a new objective function as follows:
\begin{equation} \label{vari-r}
\E_{q(R_a|\mathbf{x})} \left[\log p(w_o|r_a,w_t) \right] + H(q(R_a|\mathbf{x})) \;.
\end{equation}

\subsubsection{Approximation}
 
The partition function of Eq. (\ref{sp}) requires the summation over all opinion words in the vocabulary. 
However, the size of opinion word vocabulary is large, so we use the negative sampling technique \cite{mikolov2013distributed} to approximate Eq. (\ref{sp}).
Specifically, we substitute $\log p(w_o|r_a,w_t)$ in the objective (\ref{vari-r}) with the following objective function:
\begin{equation}
\log \sigma \big( \varphi ({w_o}, {w_t}, r_a) \big) + \sum_{ w_o' \in \NM } \log \sigma \big( - \varphi ({w_o'}, {w_t}, r_a) \big) \; ,
\end{equation}
where $w_o'$ is a negative sample of opinion words in the vocabulary, $\NM$ is the set of negative samples and $\sigma$ is the sigmoid function.
Then our final objective function is rewritten as:
\begin{equation} \label{final-o}
\begin{split}
& \E_{q(R_a|\mathbf{x})} \big[\log \sigma \big( \varphi ({w_o}, {w_t}, r_a) \big) \\
& + \sum_{ w_o' \in \NM } \log \sigma \big( - \varphi ({w_o'}, {w_t}, r_a) \big) \big] + \alpha H(q(R_a|\mathbf{x})) \;,
\end{split}
\end{equation}
where $\alpha$ is a hyper-parameter which can adjust the expectation and entropy terms into the same scale~\cite{marcheggiani2016discrete}.

\section{Target Opinion Word Pairs Extraction}
Target-opinion word pairs extraction is a well studied problem~\cite{hu2004mining,popescu2007extracting,bloom2007extracting,qiu2011opinion}. We designed five rules to extract potential target-opinion word pairs. Our method relies on Stanford Dependency Parser \cite{chen2014fast}. We describe our rules as follows.

\textbf{Rule 1}: We extract pairs satisfying the grammatical relation \textit{amod} (adjectival modifier)~\cite{de2008stanford}. For example, in phrase ``very good price,'' we extract ``price'' and ``good'' as a target-opinion pair.

\textbf{Rule 2}: We extract pairs satisfying the grammatical relation \textit{nsubj} (nominal subject), and the head word is an adjective and the tail word is a noun. For example, in a sentence ``The room is small,'' we can extract ``room'' and ``small'' as a target-opinion pair.

\textbf{Rule 3}: Some verbs are also opinion words and they are informative. We extract pairs satisfying the grammatical relation \textit{dobj} (direct object) when the head word is one of the following four words: ``like'', ``dislike'', ``love'', and ``hate''. For example, in the sentence ``I like the smell,'' we can extract ``smell'' and ``like'' as a target-opinion pair.

\textbf{Rule 4}: We extract pairs satisfying the grammatical relation \textit{xcomp} (open clausal complement), and the head word is one of the following word: ``seem'',``look'', ``feel'', ``smell'', and ``taste''. For example, in the sentence ``This beer tastes spicy,'' we can extract ``taste'' and ``spicy'' as a target-opinion pair.

\textbf{Rule 5}: If the sentence contains some adjectives that can implicitly indicate aspects, we manually assign them to the corresponding aspects. According to \cite{lakkaraju2014aspect}, some adjectives serve both as target words and opinion words. For example, in the sentence ``very tasty, and drinkable,'' the previous rules fail to extract any pair. But we know it contains a target-opinion pair, i.e., ``taste-tasty.'' 
Most of these adjectives have the same root form with the aspects they indicated, e.g., ``clean'' (cleanliness), and ``overpriced'' (price).  This kind of adjective can be extracted first and then we can obtain more similar adjectives using word similarities. For example, given ``tasty,'' we could get ``flavorful'' by retrieving similar words.

\begin{table}[tp]
\centering
\begin{tabular}{@{}c|l|l@{}}
\toprule
\multicolumn{1}{c|}{Dataset} & \multicolumn{1}{c|}{TripAdvisor} & \multicolumn{1}{c}{BeerAdvocate}  \\ \midrule
\# docs & 28,543 & 27,583 \\
\# target words & 3,737 & 3,088 \\
\# opinion words & 12,406 & 9,166 \\
\# pairs from R1 &208,676               & 249,264     \\
\# pairs from R2 &82,944             & 28,505    \\
\# pairs from R3 &2,241     & 1,092    \\
\# pairs from R4 &2,699      & 6,812    \\
\# pairs from R5 &16,537      & 55,825 \\ 
\bottomrule
\end{tabular}
\vspace{-0.1in}
\caption{Statistics of extracted target-opinion pairs .}\label{table:as_pair}
\end{table}

Table \ref{table:as_pair} shows the statistics of the rule-based extraction on our two datasets. 
The first four rules can be applied to any dataset while the last one is domain dependent which requires human effort to identify these special adjectives. 
In practice, rule 5 can be removed to save human effort. 
The effect of removing rule 5 is shown in experiments.

After extracting potential target-opinion word pairs, we need to assign them to different aspects as supervision signals. We select some seed words to describe each aspect, and then calculate similarities between the extracted target (or opinion) word and seed words, and assign the pair to the aspect where one of its seed words has the highest similarity. 
The similarity we used is the cosine similarity between two word embeddings trained by word2vec \cite{mikolov2013distributed}. 
For example, suppose seed words $\{$``room'', ``bed''$\}$ and $\{$``business'', ``Internet''$\}$ are used to describe the aspect \textit{room} and \textit{business} respectively, and the candidate pair ``pillow - soft'' will be assigned to the aspect \textit{room} if the similarity between ``pillow'' and ``bed'' is highest among all combinations. 
\section{Experiment}
In this section, we report average sentiment classification accuracies over all aspects on binary DMSC task. 
\subsection{Datasets}
We evaluate our model on TripAdvisor \cite{wang2010latent} and BeerAdvocate \cite{mcauley2012learning,lei2016rationalizing,yin2017document} datasets, which contain seven aspects (value, room, location, cleanliness, check in/front desk, service, and business) and four aspects (feel, look, smell, and taste) respectively.
We run the same preprocessing steps as \cite{yin2017document}.
Both datasets are split into train/development/test sets with proportions 8:1:1.
All methods can use development set to tune their hyper-parameters.
Ratings of TripAdvisor and BeerAdvocate datasets are on scales of $1$ to $5$ and $0$ to $5$ respectively.
But in BeerAdvocate, $0$ star is rare, so we treat the scale as $1$ to $5$.
We convert original scales to binary scales as follows: 1 and 2 stars are treated as negative, 3 is ignored, and 4 and 5 stars are treated as positive.
In BeerAdvocate, most reviews have positive polarities, so to avoid the unbalanced issue, we perform data selection according to overall polarities.
After data selection, the number of reviews with negative overall polarities and that with positive overall polarities are equal.


\begin{table*}[t]
    \centering
        \begin{tabular}{l|cccc|cccc}
            \toprule
            {\textbf{Dataset}} & \multicolumn{4}{c}{\textbf{TripAdvisor}} & \multicolumn{4}{|c}{\textbf{BeerAdvocate}} \\
            
            &  \multicolumn{2}{c}{DEV}  & \multicolumn{2}{c}{TEST}  & \multicolumn{2}{|c}{DEV}  & \multicolumn{2}{c}{TEST} \\
            & Mean & Std & Mean & Std & Mean & Std & Mean & Std \\
            \midrule
            Majority & 0.6286 &-- &0.6242&--&0.6739&--&0.6726&--\\
            Lexicon-R & 0.5914 & 0.0021 &    0.5973 & 0.0018& 0.5895&0.0020&    0.5881 & 0.0025 \\
            Lexicon-O &  0.7153 & 0.0012 & 0.7153 &0.0015 & 0.6510 & 0.0023& 0.6510 & 0.0021 \\
            Assign-O &0.7135 &0.0016 &    0.7043 & 0.0020 & 0.6652 & 0.0028 &    0.6570 & 0.0034 \\
            \hline
            N-DMSC-O & 0.7091  &--& 0.7064 & --& 0.6386 &-- &0.6493&--\\
            LRR & 0.6915 &0.0045 & 0.6947 & 0.0024 & 0.5976 &0.0110&0.5941 &0.0113\\
            \textbf{VWS-DMSC (Our)} &0.7577    & 0.0016 & 0.7561 & 0.0012 & 0.7502 & 0.0058& 0.7538 & 0.0066 \\
            \hline
            N-DMSC-50 & 0.7255    & 0.0231 & 0.7270 & 0.0204 & 0.7381 & 0.0143 & 0.7442 & 0.0157 \\
            N-DMSC-100 & 0.7482 & 0.0083 & 0.7487 & 0.0069 & 0.7443    & 0.0126 & 0.7493 & 0.0145 \\
            N-DMSC-200 & 0.7531    & 0.0040 & 0.7550 & 0.0043 &  0.7555    & 0.0096  & 0.7596 & 0.0092 \\
            N-DMSC-500 & 0.7604 & 0.0028  & 0.7616 & 0.0040 & 0.7657 & 0.0066 & 0.7713 & 0.0070 \\
            N-DMSC-1000 & 0.7631 & 0.0054 & 0.7638 & 0.0042 & 0.7708    & 0.0066  & 0.7787 & 0.0053 \\
            N-DMSC-A & 0.8281 & -- & 0.8334 & -- &0.8576 & -- & 0.8635 & --\\
            BoW-DMSC-A & 0.8027    & --&    0.8029&    --    &0.8069    &    --&0.8089&    -- \\
            \bottomrule
        \end{tabular}  
\vspace{-0.1in}
\caption{Averaged accuracies on DMSC of unsupervised, weakly supervised, and supervised methods on TripAdvisor and BeerAdvocate. Methods involve randomness also report standard deviation. }\label{table:comparison}
\vspace{-0.1in}
\end{table*}

\subsection{Compared Methods}

To demonstrate the effectiveness of our method, we compare our model with following baselines:

\textbf{Majority} uses the majority of sentiment polarities in
training sets as predictions.

\textbf{Lexicon} means using an opinion lexicon to assign sentiment polarity to an aspect \cite{read2009weakly,pablos2015v3}.
We combine two popular opinion lexicons used by \citet{hu2004mining} and \citet{wilson2005recognizing} to get a new one.
If an opinion word from extracted pairs is in positive (negative) lexicon, it votes for positive (negative).
When the opinion word is with a negation word, its polarity will be flipped.
Then, the polarity of an aspect is determined by using majority voting among all opinion words associated with the aspect.
When the number of positive and negative words is equal, we adopt two different ways to resolve it.
For \textbf{Lexicon-R}, it randomly assigns a polarity.
For \textbf{Lexicon-O}, it uses the overall polarity as the prediction.
Since overall polarities can also be missing, for both Lexicon-R and Lexicon-O, we randomly assign a polarity in uncertain cases and report both mean and std based on five trials of random assignments. 

\textbf{Assign-O} means directly using the overall polarity of a review in the development/test sets as the prediction for each aspect.

\textbf{LRR} assumes the overall polarity is a weighted sum of the polarity of each aspect \cite{wang2010latent}.
LRR can be regarded as the only existing weakly supervised baseline where both algorithm and source code are available.

\textbf{BoW-DMSC-A} is a simple softmax classifier using all annotated training data where the input is a bag-of-words feature vector of a document.  

\textbf{N-DMSC-A} is the state-of-the-art neural network based model~\cite{yin2017document} (\textbf{N-DMSC}) in DMSC task using all annotated training data, which serves an upper bound to our method.

\textbf{N-DMSC-O} is to use overall polarities as ``supervision'' to train an N-DMSC and apply it to the classification task of each aspect at the test time. 

\textbf{N-DMSC-\{50,100,200,500,1000\}} is the N-DMSC algorithm using partial data. 
In order to see our method is comparable to supervised methods using how many labeled data, we use $\{50,100,200,500,1000\}$ annotations of each aspect to train N-DMSC and compare them to our method.
In addition to annotated data for training, there are extra $20\%$ annotated data for validation.
Since the sampled labeled data may vary for different trials, we perform five trials of random sampling and report both mean and std of the results.

For our method, denoted as \textbf{VWS-DMSC}, the document representation we used is obtained from N-DMSC \cite{yin2017document}. 
They proposed a novel hierarchical iterative attention model in which documents and pseudo aspect related questions are interleaved at both word and sentence-level to learn an aspect-aware document representation. The pseudo aspect related questions are represented by aspect related keywords.
In order to benefit from their aspect-aware representation scheme, we train an N-DMSC to extract the document representation using only overall polarities. 
In the iterative attention module, we use the pseudo aspect related keywords of all aspects released by \citet{yin2017document}. 
One can also use document-to-document autoencoders \cite{li2015hierarchical} to generate the document representation.
In this way, our method can get rid of using overall polarities to generate the document representation. 
Hence, unlike LRR, it is not necessary for our method to use overall polarities. 
Here, to have a fair comparison with LRR, we use the overall polarities to generate document representation. 
For our method, we do not know which state is positive and which one is negative at training time, so the Hungarian algorithm \cite{kuhn1955hungarian} is used to resolve the assignment problem at the test time. 


\subsection{Results and Analysis}
We show all results in Table~\ref{table:comparison}, which consists of three blocks, namely, unsupervised, weakly supervised, and supervised methods. 

For unsupervised methods, our method can outperform majority on both datasets consistently. 
But other weakly supervised methods cannot outperform majority on BeerAdvocate dataset, which shows these baselines cannot handle unbalanced data well since BeerAdvocate is more unbalanced than TripAdvisor. 
Our method outperforms Lexicon-R and Lexicon-O, which shows that predicting an opinion word based on a target word may be a better way to use target-opinion pairs, compared with performing a lexicon lookup using opinion words from extract pairs.
Good performance of Lexicon-O and Assign-O demonstrates the usefulness of overall polarities in development/test sets. 
N-DMSC-O trained with the overall polarities cannot outperform Assign-O since N-DMSC-O can only see overall polarities in training set while Assign-O can see overall polarities for both development and test sets and does not involve learning and generalization. 

For weakly supervised methods, LRR is the only open-source baseline in the literature on weakly supervised DMSC, and our method outperforms LRR by \textbf{6}\% and \textbf{16}\% on TripAdvisor and BeerAdvocate datasets. 
N-DMSC-O can also be considered as a weakly supervised method because it only uses overall polarities as ``supervision,'' and we still outperform it significantly. 
It is interesting that LRR is worse than N-DMSC-O. 
We guess that assuming that the overall polarity is a weighted sum of all aspect polarities may not be a good strategy to train each aspect's polarity or the document representation learned by N-DMSC is better than the bag-of-words representation.

For supervised block methods, BoW-DMSC-A and N-DMSC-A are both supervised methods using all annotated data, which can be seen as the upper bound of our algorithm.
N-DMSC-A outperforms BoW-DMSC-A, which shows that the document representation based on neural network is better than the bag-of-words representation. 
Hence, we use the neural networks based document representation as input of the sentiment polarity classifier. 
Our results are comparable to N-DMSC-200 on TripAdvisor and N-DMSC-100 on BeerAdvocate. 

\begin{table}[t]
    \centering
        \begin{tabular}{c|cc|cc}
            \toprule
            {\textbf{Dataset}} & \multicolumn{2}{c}{\textbf{TripAdvisor}} & \multicolumn{2}{|c}{\textbf{BeerAdvocate}} \\
            Rule & DEV  & TEST  & DEV  & TEST \\
            \midrule
            R1 & 0.7215 & 0.7174 & 0.7220 & 0.7216 \\
            R2 & 0.7172 & 0.7180 & 0.6864 & 0.6936 \\
            R3 & 0.6263 & 0.6187 & 0.6731 & 0.6725 \\
            R4 & 0.6248 & 0.6279 & 0.6724 & 0.6717 \\
            R5 & 0.5902 & 0.5856 & 0.7095 & 0.7066 \\
            - R1 & 0.7538 & 0.7481 & 0.7458 & 0.7474 \\
            - R2 & 0.7342 & 0.7368 & 0.7504 & 0.7529 \\
            - R3 & 0.7418 & 0.7397 & 0.7565 & 0.7558 \\
            - R4 & 0.7424 & 0.7368 & 0.7518 & 0.7507 \\
            - R5 & 0.7448 & 0.7440 & 0.7550 & 0.7548 \\
            \hline
            All & 0.7577 & 0.7561 & 0.7502 & 0.7538 \\
            \bottomrule
        \end{tabular}
\vspace{-0.1in}
\caption{Averaged accuracies on DMSC. ``R1\;--\;R5'' means only using a rule while ``-R1\;--$\;$-R5'' means removing a rule from all the rules.}  \label{table:ablation}
\vspace{-0.1in} 
\end{table}

\subsection{Ablation Study}
To evaluate effects of extracted rules, we performed an ablation study.
We run our algorithm VWS-DMS with each rule kept or removed over two datasets. 
If no pairs extracted for one aspect in training set, the accuracy of this aspect will be 0.5, which is a random guess. 
From the Table~\ref{table:ablation} we can see that, the rule R1 is the most effective rule for both datasets. Rules R3/R4/R5 are less effective on their own. 
However, as a whole, they can still improve the overall performance.
When considering removing each of rules, we found that our algorithm is quite robust, which indicates missing one of the rules may not hurt the performance much. 
Hence, if human labor is a major concern, rule 5 can be discarded.
We found that sometimes removing one rule may even result in better accuracy (e.g., ``-R3'' for BeerAdvocate dataset).
This means this rule may introduce some noises into the objective function.
However, ``-R3'' can result in worse accuracy for TripAdvisor, which means it is still complementary to the other rules for this dataset.

\subsection{Parameter Sensitivity}
\begin{figure}[t]
    \includegraphics[width=0.5\textwidth]{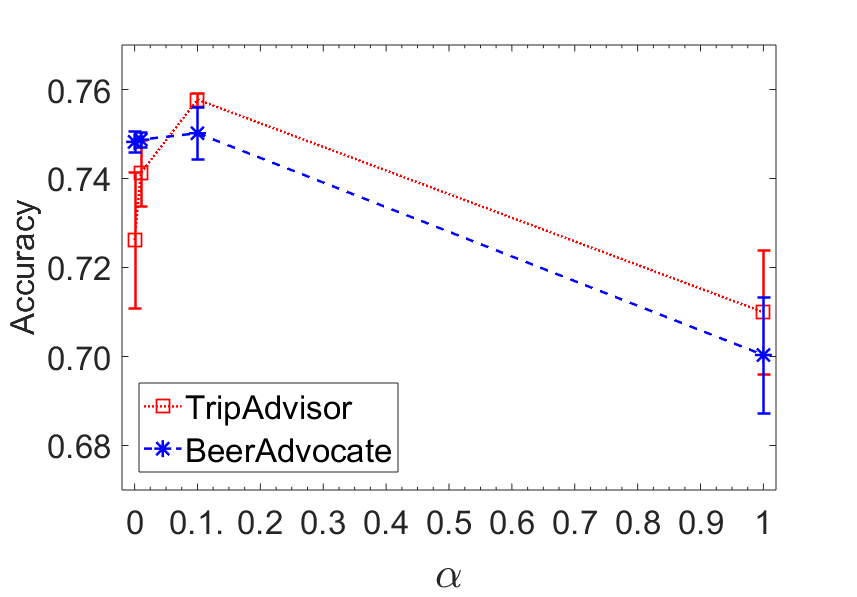}\vspace{-0.05in}
    \vspace{-0.2in}
    \caption{Parameter sensitivity analysis.}\label{fig:alpha}
\end{figure}

We also conduct parameter sensitivity analysis of our approach. The parameter $\alpha$ in Equation~(\ref{final-o}) adjusts the expectation and entropy terms on the same scale. We test $\alpha=\{0, 0.01,0.1,1\}$ for both of the datasets.
As we can see from Figure~\ref{fig:alpha}, $\alpha=0.1$ is a good choice for both datasets.

\subsection{Implementation Details}
We implemented our models using TensorFlow \cite{abadi2016}.
For N-DMSC and LRR, we used code released by \citet{yin2017document} and \citet{wang2010latent} respectively and followed their preprocessing steps and optimal settings.

Parameters are updated by using ADADELTA \cite{zeiler2012adadelta}, an adaptive learning rate method.
To avoid overfitting, we impose weight decay and drop out on both classifiers.
The regularization coefficient and drop out rate are set to $10^{-3}$ and $0.3$ respectively.
The number of negative samples and $\alpha$ in our model are set to $10$ and $0.1$ respectively.
For each document and each aspect, multiple target-opinion pairs are extracted.
The opinion word classifier associated with an aspect will predict five target-opinion pairs at a time.
These five target-opinion pairs are selected with bias.
The probability of a pair being selected is proportional to the frequency of the opinion word to the power of $-0.25$.
In this way, opinion words with low frequency are more likely to be selected compared to the uniform sampling. 
In order to initialize both classifiers better, the word embeddings are retrofitted \cite{faruqui2015} using PPDB \cite{ganitkevitch2013ppdb} semantic lexicons. 
\section{Related Work}

In this section, we review the related work on document-level multi-aspect sentiment classification, target-opinion word pairs extraction, and variational methods.

\paragraph{Document-level Multi-Aspect Sentiment Classification.}
\citet{wang2010latent} proposed a LRR model to solve this problem.
LRR assumes the overall polarity is a weighted sum of all aspect polarities which are represented by word frequency features.
LRR needs to use aspect keywords to perform sentence segmentation to generate the representation of each aspect.
To address the limitation of using aspect keywords, LARAM \cite{wang2011latent} assumes that the text content describing a particular aspect is generated by sampling words from a topic model corresponding to the latent aspect.
Both LRR and LARAM can only access to overall polarities in the training data, but not gold standards of aspect polarities.
\citet{meng2018weakly} proposed a weakly supervised text classification method which can take label surface names, class-related keywords, or a few labeled documents as supervision. 
\citet{ramesh2015weakly} developed a weakly supervised joint model to identify aspects and the corresponding sentiment polarities in online courses. They treat aspect (sentiment) related seed words as weak supervision. 
In the DMSC task which is a fine-grained text classification task, the label surface names or keywords for some aspects would be very similar. 
Given that the inputs are the same and the supervisions are similar, weakly supervised models cannot distinguish them. 
So we do not consider them as our baselines. 
\citet{yin2017document} modeled this problem as a machine comprehension problem under a multi-task learning framework.
It also needs aspect keywords to generate aspect-aware document representations.
Moreover, it can access gold standards of aspect polarities and achieved state-of-the-art performance on this task.
Hence, it can serve as an upper bound.
Some sentence-level aspect based sentiment classification methods \cite{wang2016attention, wang2018aspect} can be directly applied to the DMSC task, because they can solve aspect category sentiment classification task. 
For example, given a sentence ``the restaurant is expensive,'' the aspect category sentiment classification task aims to classify the polarity of the aspect category ``price'' to be \textit{negative}. 
The aspect categories are predefined which are the same as the DMSC task. 
Some of them \cite{tang2016effective, tang2016aspect, chen2017recurrent, ma2017interactive} cannot because they are originally designed for aspect term sentiment classification task. 
For example, given a sentence ``I loved their fajitas,'' the aspect term sentiment classification task aims to classify the polarity of the aspect term ``fajitas'' to be \textit{positive}.  
The aspect terms appearing in the sentence should be provided as inputs. 

\paragraph{Target Opinion Word Pairs Extraction.}
There are two kinds of methods, namely, rule based methods and learning based methods to solve this task. 
Rule based methods extract target-opinion word pairs by mining the dependency tree paths between target words and opinion words. 
Learning based methods treat this task as a sequence labeling problem, mapping each word to one of the following categories: target, opinion, and other.

\cite{hu2004mining} is one of earliest rule based methods to extract target-opinion pairs. An opinion word is restricted to be an adjective. Target words are extracted first, and then an opinion word is linked to its nearest target word to form a pair.
\citet{popescu2007extracting} and \citet{bloom2007extracting} manually designed dependency tree path templates to extract target-opinion pairs. If the path between a target word candidate and an opinion word candidate belongs to the set of path templates, the pair will be extracted. 
\citet{qiu2011opinion} identified dependency paths that link opinion words and targets via a bootstrapping process. 
This method only needs an initial opinion lexicon to start the bootstrapping process. 
\citet{zhuang2006movie} adopted a supervised learning algorithm to learn valid dependency tree path templates, but it requires target-opinion pairs annotations. 

Learning based methods require lots of target-opinion pairs annotations. 
They trained conditional random fields (CRF) \cite{lafferty2001conditional} based models \cite{jakob2010extracting, yang2012extracting, wang2016recursive} or deep neural networks \cite{liu2015fine,wenya2017coupled,li2017deep} to predict the label (target, opinion or other) of each word. 
\citet{jakob2010extracting} and \citet{li2012cross} extracted target-opinion pairs without using using any labeled data in the domain of interest, but it needs lots of labeled data in another related domain.

In this paper, we only use very simple rules to extract target-opinion pairs to validate the effectiveness of our approach. 
If better pairs can be extracted, we can further improve our results.

\paragraph{Variational Methods.}
Variational autoencoders~\cite{KingmaW13,rezende2014stochastic} (VAEs) use a neural network to parameterize a probability distribution. 
VAEs consists of an encoder which parameterizes posterior probabilities and a decoder which parameterizes the reconstruction likelihood given a latent variable. 
VAEs inspire many interesting works \cite{titov2014unsupervised,marcheggiani2016discrete,vsuster2016bilingual,zhang2018variational, chen2018variational} which are slightly different from VAEs. Their encoders produce a discrete distribution while the encoder in VAEs yields a continuous latent variable. 
\citet{titov2014unsupervised} aimed to solve semantic role labeling problem. The encoder is essentially a semantic role labeling model which predicts roles given a rich set of syntactic and lexical features. The decoder reconstructs argument fillers given predicted roles.
\citet{marcheggiani2016discrete} aimed to solve unsupervised open domain relation discovery. The encoder is a feature-rich relation extractor, which predicts a semantic relation between two entities. The decoder reconstructs entities relying on the predicted relation.
\citet{vsuster2016bilingual} tried to learn multi-sense word embeddings. The encoder uses bilingual context to choose a sense for a given word. The decoder predicts context words based on the chosen sense and the given word. 
\citet{zhang2018variational} aimed to solve knowledge graph powered question answering. 
Three neural networks are used to parameterize probabilities of a topic entity given a query and an answer, an answer based on a query and a predicted topic, and the topic given the query. 
\citet{chen2018variational} aimed to infer missing links in a knowledge graph. 
Three neural networks are used to parameterize probabilities of a latent path given two entities and a relation, a relation based on two entities and the chosen latent path, and the relation given the latent path. 
Our method also uses neural networks to parameterize two discrete distributions but aims to solve the DMSC task. 
\section{Conclusion}
In this paper, we propose a variational approach to weakly supervised DMSC.
We extract many target-opinion word pairs from dependency parsers using simple rules.
These pairs can be ``supervision'' signals to predict sentiment polarity. 
Our objective function is to predict an opinion word given a target word. 
After introducing the sentiment polarity as a latent variable, we can learn a sentiment polarity classifier by optimizing the variational lower bound. 
We show that we can outperform weakly supervised baselines by a large margin and achieve comparable results to the supervised method with hundreds of labels per aspect, which can reduce a lot of labor work in practice.
In the future, we plan to explore better target-opinion word extraction approaches to find better ``supervision'' signals.

\section*{Acknowledgments}
This paper was supported by the Early Career Scheme (ECS, No. 26206717) from Research Grants Council in Hong Kong. Ziqian Zeng has been supported by the Hong Kong Ph.D. Fellowship. We thank Intel Corporation for supporting our deep learning related research. 
We also thank the anonymous reviewers for their valuable comments and suggestions that help improve the quality of this manuscript.

\bibliography{naaclhlt2019}
\bibliographystyle{acl_natbib}

\end{document}